%% file: main.tex
\documentclass{article}

\PassOptionsToPackage{numbers, compress}{natbib}
\usepackage[dblblindworkshop, final]{neurips_2025}

\workshoptitle{Math-AI}

\usepackage{caption}
\usepackage{subcaption}
\usepackage[utf8]{inputenc} % allow utf-8 input
\usepackage[T1]{fontenc}    % use 8-bit T1 fonts
\usepackage{hyperref}       % hyperlinks
\usepackage{url}            % simple URL typesetting
\usepackage{booktabs}       % professional-quality tables
\usepackage{amsfonts}       % blackboard math symbols
\usepackage{nicefrac}       % compact symbols for 1/2, etc.
\usepackage{microtype}      % microtypography
\usepackage{xcolor}         % colors
\usepackage{graphicx}
\usepackage{amsmath}
\usepackage{amsmath,amsthm,amssymb}
\usepackage{mathtools}
\usepackage{tcolorbox}
\usepackage{wrapfig}
\title{Towards Understanding \\Self-play for LLM Reasoning}

\author{%
  Justin Yang Chae\\%\thanks{Use footnote for providing further information
  University of Washington\\
  Seattle, WA, USA \\
  \texttt{jchae3@uw.edu}
  \And
  Md Tanvirul Alam \\
  Rochester Institute of Technology \\
  Rochester, NY, USA \\
  \texttt{ma8235@rit.edu}
  \AND
  Nidhi Rastogi \\
  Rochester Institute of Technology \\
  Rochester, NY, USA \\
  \texttt{nxrvse@rit.edu}
}

\newtheorem{theorem}{Theorem}
\newtheorem{corollary}{Corollary}

\begin{document}

\maketitle

\input{sections/abstract}

\input{sections/intro}
\input{sections/background}

\section{Experiments}
We adopt the \textbf{Absolute Zero Reasoner} (AZR)~\citep{zhao2025absolute} codebase as our self-play framework. Due to computational constraints, we only conduct experiments on \textsc{Qwen2.5-Coder-3B} and \textsc{Qwen2.5-Coder-7B}.

We investigate the following research questions:

\begin{description}
    \item[RQ1:] Does self-play in AZR enable novel reasoning beyond the capacity of the base model?
    \item[RQ2:] How does the difficulty of the questions proposed evolve over the course of training?
    \item[RQ3:] Does AZR experience entropy collapse during training like RLVR-trained models?
    \item[RQ4:] How does parameter update sparsity in self-play compare with RLVR and SFT?
    \item[RQ5:] Does lowering the proposer's target question difficulty improve model performance?
\end{description}

\input{sections/pass_k}
\input{sections/question}

\input{sections/entropy}

\input{sections/sparsity}

\input{sections/reward}
\input{sections/conclusion}

\clearpage
\bibliographystyle{ACM-Reference-Format}
\bibliography{ref}
\clearpage
\input{appendix}

\end{document}

%% file: sections/abstract.tex
\begin{abstract}
Recent advances in large language model (LLM) reasoning, led by reinforcement learning with verifiable rewards (RLVR), have inspired self-play post-training, where models improve by generating and solving their own problems. While self-play has shown strong in-domain and out-of-domain gains, the mechanisms behind these improvements remain poorly understood. In this work, we analyze the training dynamics of self-play through the lens of the \textit{Absolute Zero Reasoner}, comparing it against RLVR and supervised fine-tuning (SFT). Our study examines parameter update sparsity, entropy dynamics of token distributions, and alternative proposer reward functions. We further connect these dynamics to reasoning performance using pass@k evaluations. Together, our findings clarify how self-play differs from other post-training strategies, highlight its inherent limitations, and point toward future directions for improving LLM math reasoning through self-play.
\end{abstract}

%% file: sections/intro.tex
\section{Introduction}

Reinforcement learning with verifiable rewards (RLVR) has emerged as the leading method for improving reasoning in large language models (LLMs), with notable successes in mathematics and other verifiable domains~\citep{shao2024deepseekmath, lambert2025tulu3pushingfrontiers}. However, its benefits remain debated: recent studies argue that RLVR primarily sharpens output distributions and improves sampling efficiency rather than fostering genuine reasoning~\citep{yue2025does, dang2025assessing}. Analyses of its training dynamics further show reduced policy entropy and sparser parameter updates relative to supervised fine-tuning~\citep{cui2025entropymechanismreinforcementlearning, mukherjee2025reinforcement}.

Beyond RLVR, recent work has investigated the potential of self-play, an alternative paradigm where models interact with themselves to drive improvement. For example, \citet{zhao2025absolute} introduced a framework where an LLM acts as both a proposer of coding problems and a solver, achieving significant gains on out-of-domain mathematics benchmarks using only its own coding outputs and without additional human-curated data. Similarly, other methods have shown the promise of self-play for LLMs~\citep{liu2025chasing, huang_r-zero_2025, liang_beyond_2025}. The success of these methods highlights a critical need to understand how models can improve without external supervision and whether these improvements reflect genuine gains in reasoning ability.

In this work, we provide an analysis of the self-play approach, comparing its training dynamics and performance against RLVR and SFT. We follow the \textit{pass$@k$} experiments of \citet{yue2025does} to rigorously assess whether self-play-trained models achieve stronger reasoning performance than their base counterparts. We study internal training dynamics through measures of policy entropy and parameter update sparsity to shed light on how self-play shapes model behavior. Finally, we investigate the role of the proposer component by modifying its reward function and we also track how the distribution and difficulty of generated questions evolve throughout training.

%% file: sections/background.tex
\section{Absolute Zero Reasoner}
% The \emph{Absolute Zero} paradigm~\cite{zhao2025absolute} introduces a pure self-play framework where a single model simultaneously proposes tasks and solves them, thereby learning from both stages. The environment validates each proposed task to form a reasoning problem with a gold label, after which the same model attempts to solve it and receives rewards for both task learnability and solution quality. This iterative training loop enables the agent to continually generate new challenges while progressively improving its reasoning and problem-solving abilities.  

% In the same work, Zhao et al.~\cite{zhao2025absolute} developed the \emph{Absolute Zero Reasoner} (AZR) as a concrete instantiation of this paradigm. AZR employs a unified LLM that acts in two roles: as a proposer, it generates tasks to expand its own learning curriculum, and as a solver, it attempts to solve them to improve its reasoning ability. The environment validates each proposed task to ensure a well-defined problem with a gold solution, and the model receives feedback on both task quality and solution accuracy. To provide structure, AZR trains on three families of coding-based tasks that capture abductive, deductive, and inductive reasoning. The model is jointly optimized across both roles using a multitask advantage estimator, allowing it to refine its curriculum, generate novel challenges, and strengthen its problem-solving capabilities over time. We discuss the implementation details of AZR in Appendix~\ref{sec:AZR}. 

The \emph{Absolute Zero} paradigm~\cite{zhao2025absolute} is a self-play framework where a single model learns by simultaneously proposing and solving tasks. One implementation, the \emph{Absolute Zero Reasoner} (AZR), uses a unified LLM that acts as both a \textbf{proposer}, which generates a curriculum of tasks, and a \textbf{solver}, which learns by solving them. After an environment validates each proposed task to create a problem with a gold solution, the model is jointly optimized using a multitask advantage estimator, receiving rewards for both task quality and solution accuracy. AZR is trained on three families of coding tasks designed to capture abductive, deductive, and inductive reasoning; further implementation details are available in Appendix~\ref{sec:AZR}.

%% file: sections/pass_k.tex
\subsection{RQ1: Reasoning Capacity}
\label{pass@k}   
% \begin{wrapfigure}{r}{0.5\textwidth}
%     \vspace{-10pt} % Optional: Adjusts vertical space if needed
%     \centering
%     % --- Top Row of Grid ---
%     \begin{subfigure}[b]{0.24\textwidth}
%         \includegraphics[width=\textwidth]{figures/3b_mbpp_pass_curves.pdf}
%     \end{subfigure}
%     \hfill % Adds flexible space between the two top figures
%     \begin{subfigure}[b]{0.24\textwidth}
%         \includegraphics[width=\textwidth]{figures/3b_humaneval_pass_curves.pdf}
%     \end{subfigure}
    
%     \vspace{3pt} % Adds a small vertical space between the rows

%     % --- Bottom Row of Grid ---
%     \begin{subfigure}[b]{0.24\textwidth}
%         \includegraphics[width=\textwidth]{figures/3b_aime24_pass_curves.pdf}
%     \end{subfigure}
%     \hfill % Adds flexible space between the two bottom figures
%     \begin{subfigure}[b]{0.24\textwidth}
%         \includegraphics[width=\textwidth]{figures/3b_aime25_pass_curves.pdf}
%     \end{subfigure}

%     \caption{\textbf{Self-play models are still bounded by the base model.} Pass@k curves for \textsc{AZR-Coder-3B} (red) and its base model \textsc{Qwen2.5-Coder-3B} (blue) across four different benchmarks.}
%     \label{fig:passk_curves}
% \end{wrapfigure}
\begin{figure}[t]
    \centering
    \begin{subfigure}[b]{0.25\textwidth}
        \includegraphics[width=\textwidth]{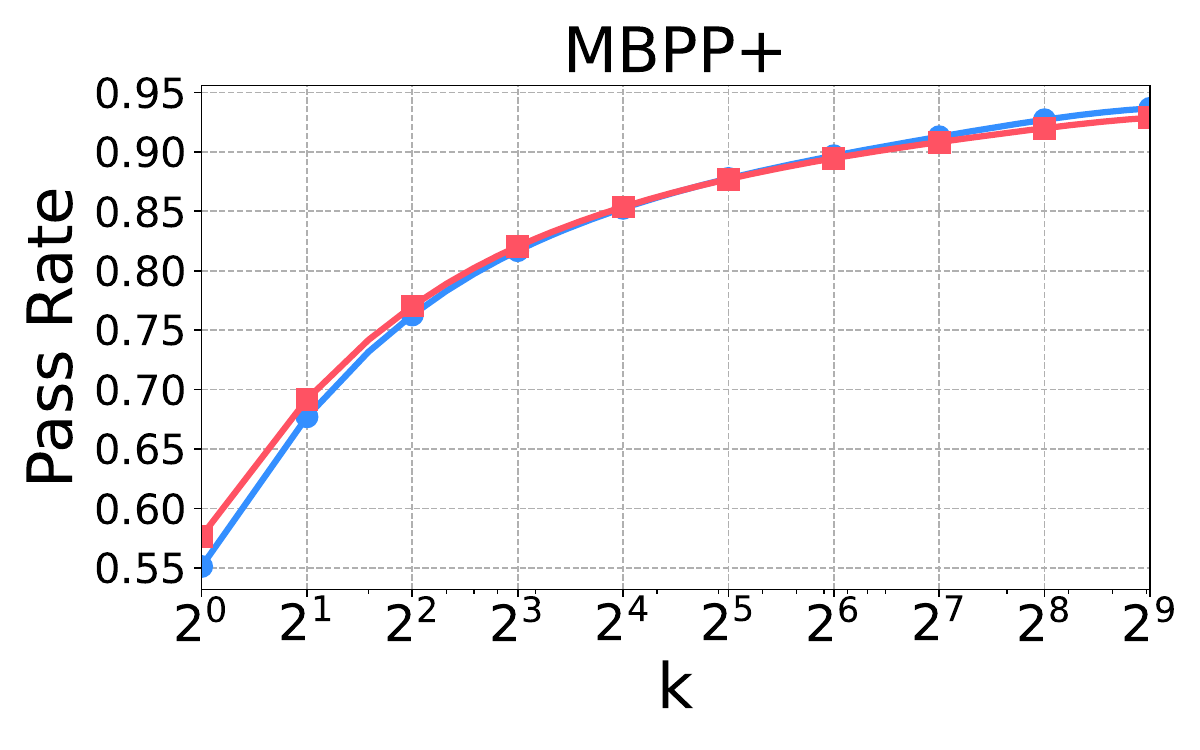}
    \end{subfigure}
    \begin{subfigure}[b]{0.24\textwidth}
        \includegraphics[width=\textwidth]{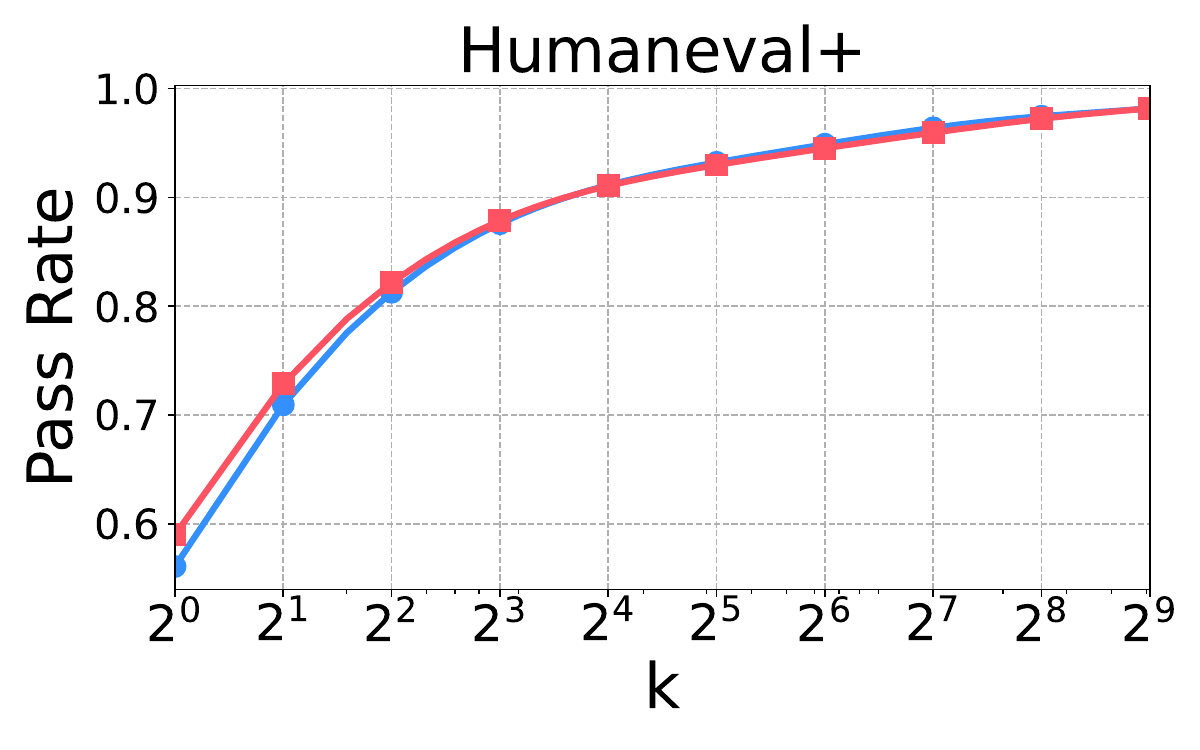}
    \end{subfigure}
    \begin{subfigure}[b]{0.24\textwidth}
        \includegraphics[width=\textwidth]{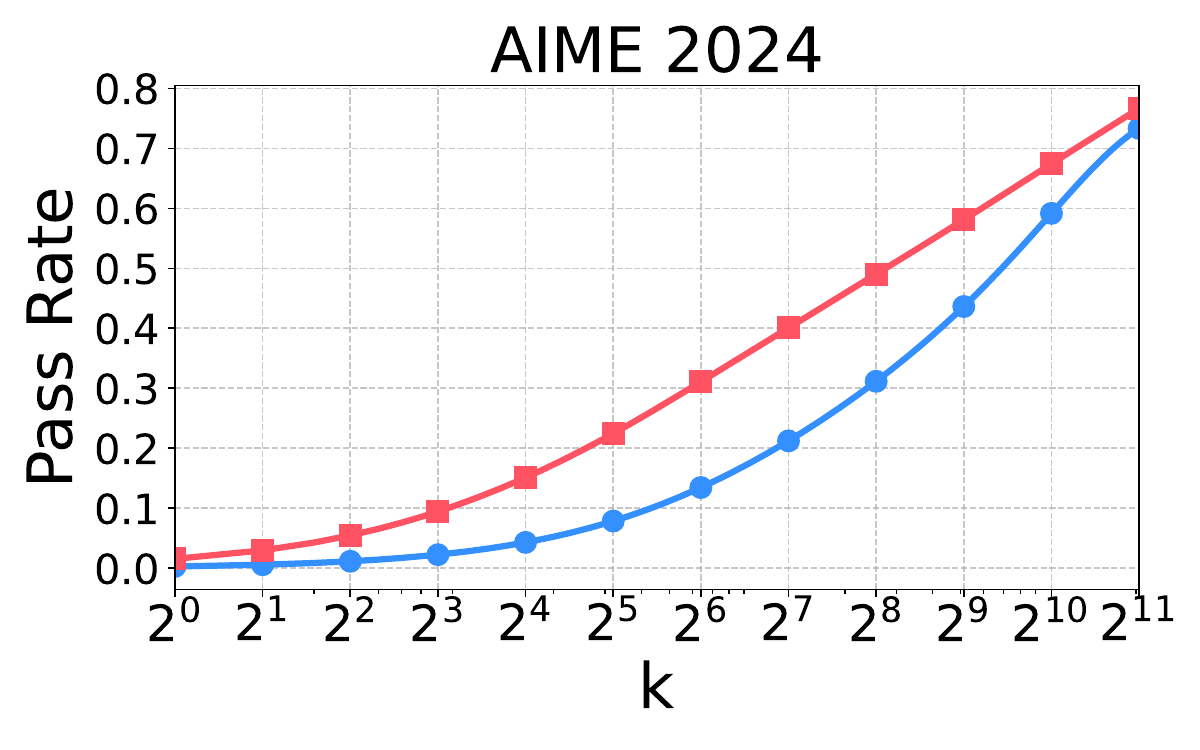}
    \end{subfigure}
    \begin{subfigure}[b]{0.25\textwidth}
        \includegraphics[width=\textwidth]{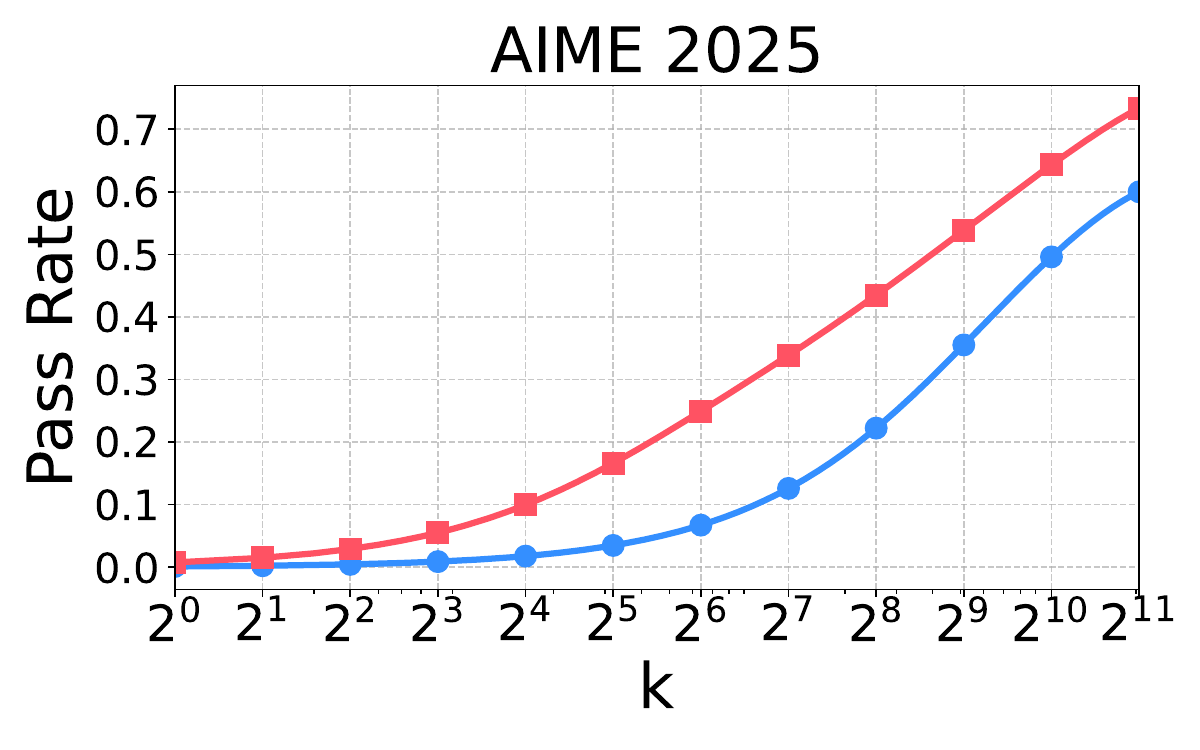}
    \end{subfigure} \\
    \begin{subfigure}[b]{0.25\textwidth}
        \includegraphics[width=\textwidth]{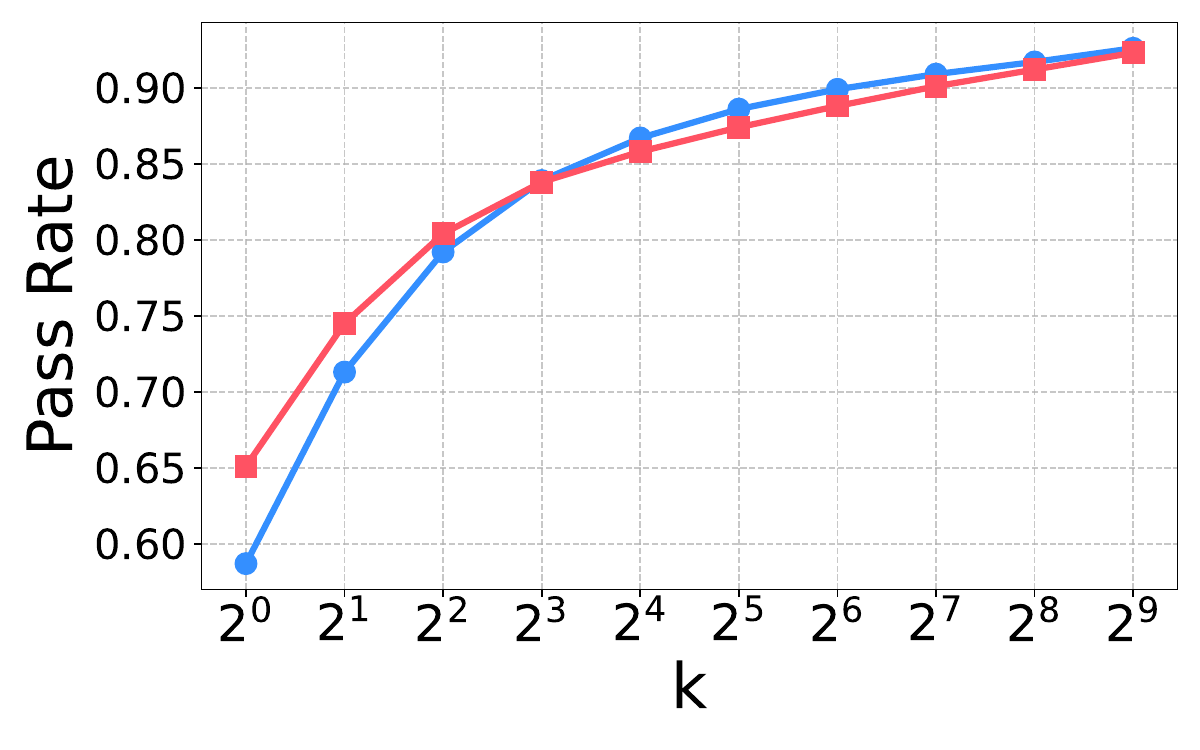}
    \end{subfigure}
    \begin{subfigure}[b]{0.24\textwidth}
        \includegraphics[width=\textwidth]{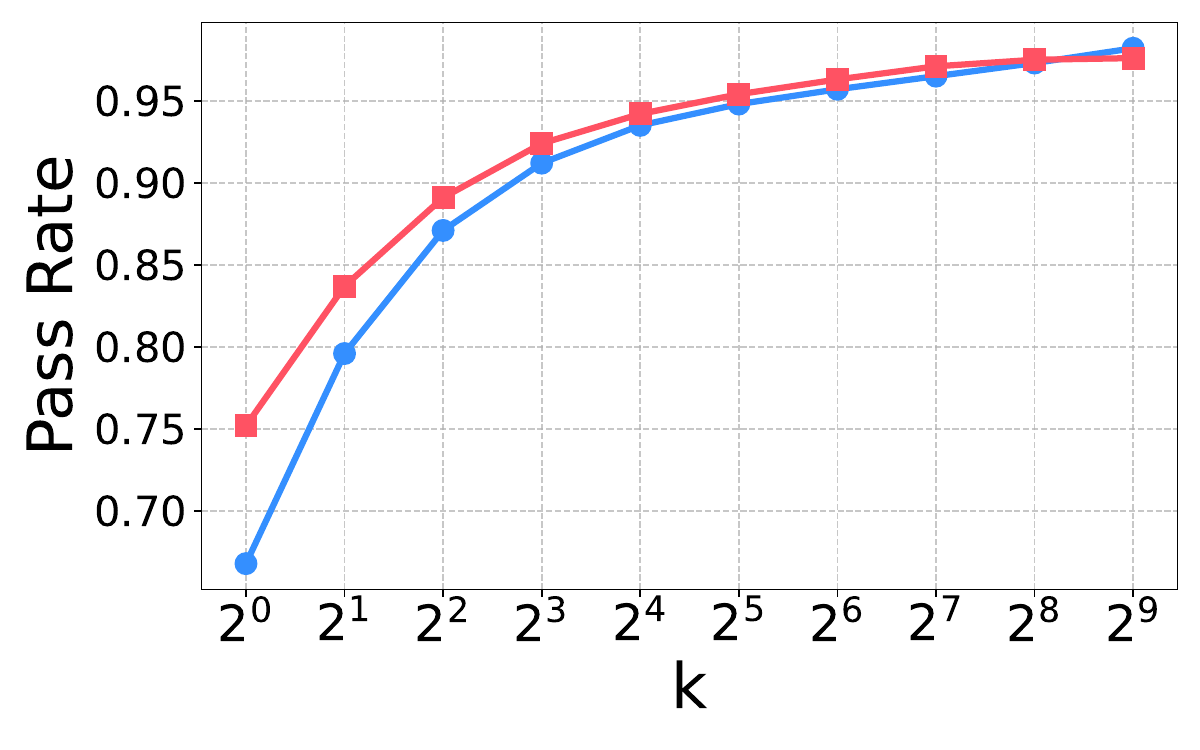}
    \end{subfigure} 
    \begin{subfigure}[b]{0.24\textwidth}
        \includegraphics[width=\textwidth]{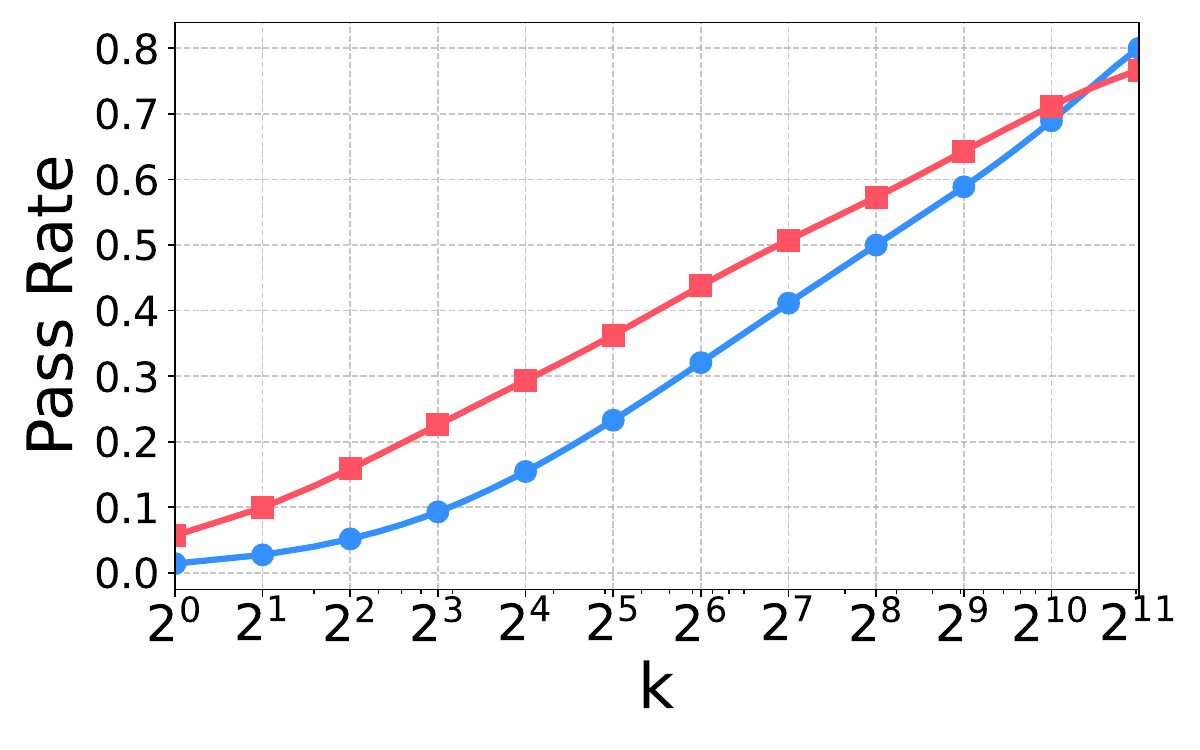}
    \end{subfigure}
    \begin{subfigure}[b]{0.25\textwidth}
        \includegraphics[width=\textwidth]{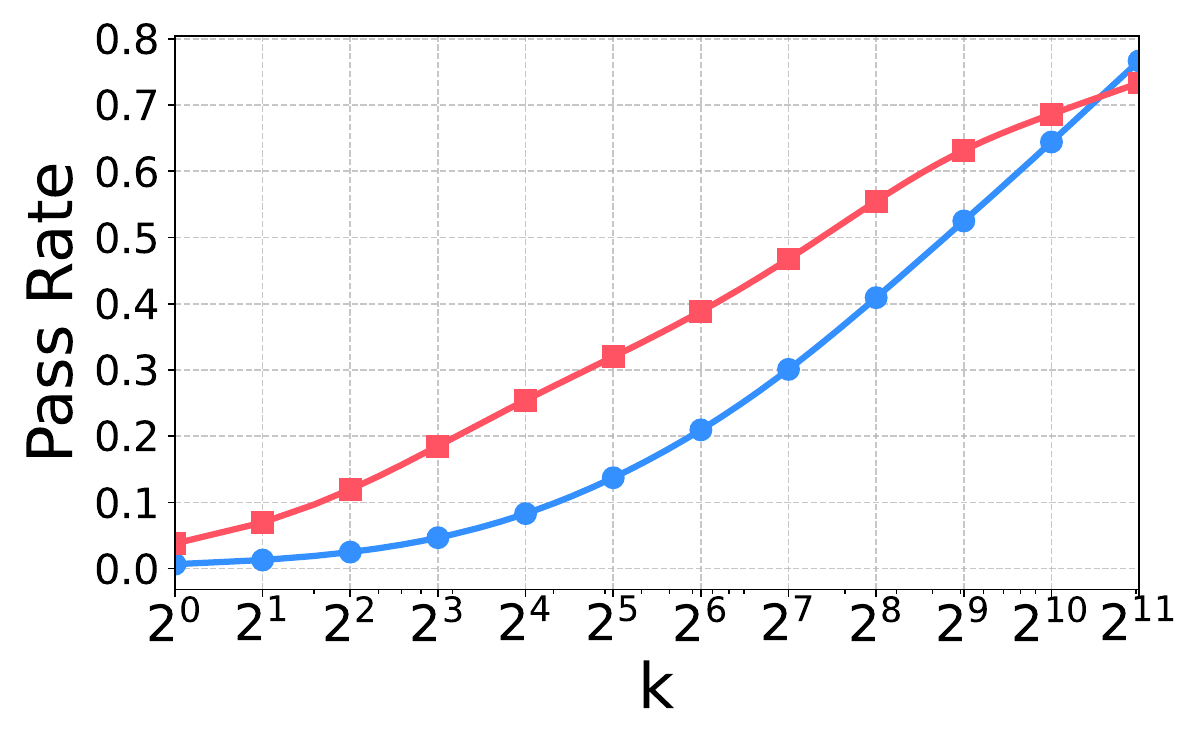}
    \end{subfigure}
    \caption{\textbf{Self-play models are still bounded by the base model.} (Top row) Pass@k curves for \textsc{AZR-Coder-3B} (red) and \textsc{Qwen2.5-Coder-3B} (blue). (Bottom row) Pass@k curves for \textsc{AZR-Coder-7B} (red) and \textsc{Qwen2.5-Coder-7B} (blue).}
    \label{fig:passk_curves} 
\end{figure}

\textbf{Setup.} We use an unbiased pass$@k$ estimator~\citep{chen2021evaluating} to measure reasoning capacity, following the experimental setup of \citet{yue2025does} on benchmarks including MBPP+, HumanEval+, and AIME. 

\textbf{Results.} We show that self-play improves performance at small k, but the base model performs better at large k (Figure~\ref{fig:passk_curves}). This suggests gains from distributional sharpening~\citep{yue2025does, dang2025assessing}. However, the performance drop at large k is not statistically significant, suggesting it preserves the base model's capacity more effectively than standard RLVR. We hypothesize this is due to implicit data diversity from the co-evolutionary training, a factor suggested to sustain pass$@k$ gains in \citet{liang_beyond_2025}.

We provide a formal argument for why AZR remains bounded by the reasoning capacity of the base model, building on the analysis of \citet{wu_invisible_2025}, in Appendix~\ref{sec:azr-il}.

\begin{tcolorbox}[colback=gray!10, colframe=black, title=Takeaway]
AZR is effectively RLVR over a proposer-induced task distribution \cite{zhao2025absolute}, and thus inherits the same Invisible-Leash support limitation \cite{wu_invisible_2025}: self-play cannot assign probability mass to solutions outside the base model’s support. As a result, AZR remains bounded by the reasoning capacity of the base model, though unlike RLVR, the drop in pass$@k$ performance at larger $k$ is less pronounced.
\end{tcolorbox}

%% file: sections/question.tex
\subsection{RQ2: Evolution of Question Difficulty}
\begin{wrapfigure}{r}{0.5\textwidth}
    \centering
    \vspace{-10pt} % Optional: Adjusts vertical spacing if needed
    \begin{subfigure}[b]{0.24\textwidth}
        \includegraphics[width=\textwidth]{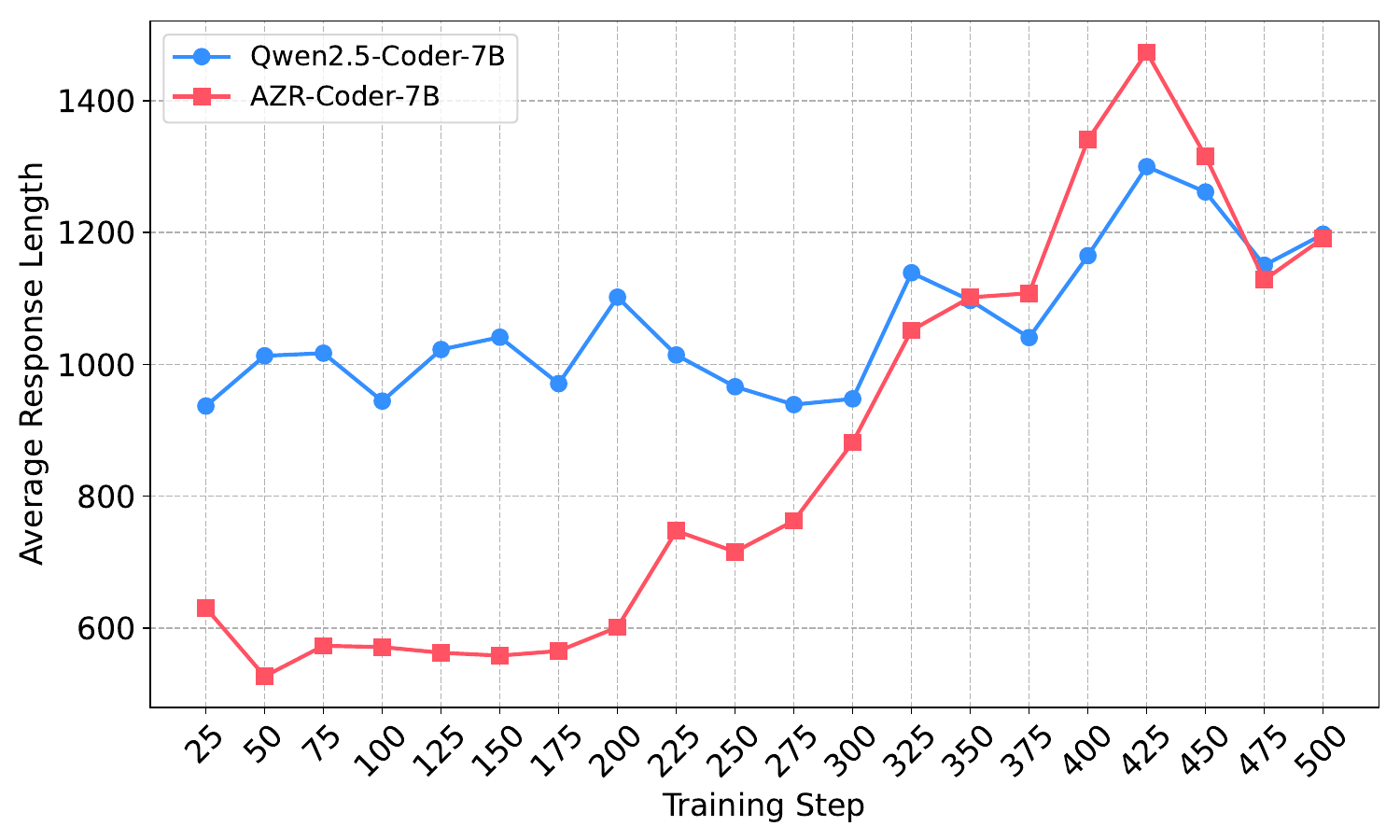}
    \end{subfigure}
    \hfill % Adds flexible space between the figures
    \begin{subfigure}[b]{0.24\textwidth}
        \includegraphics[width=\textwidth]{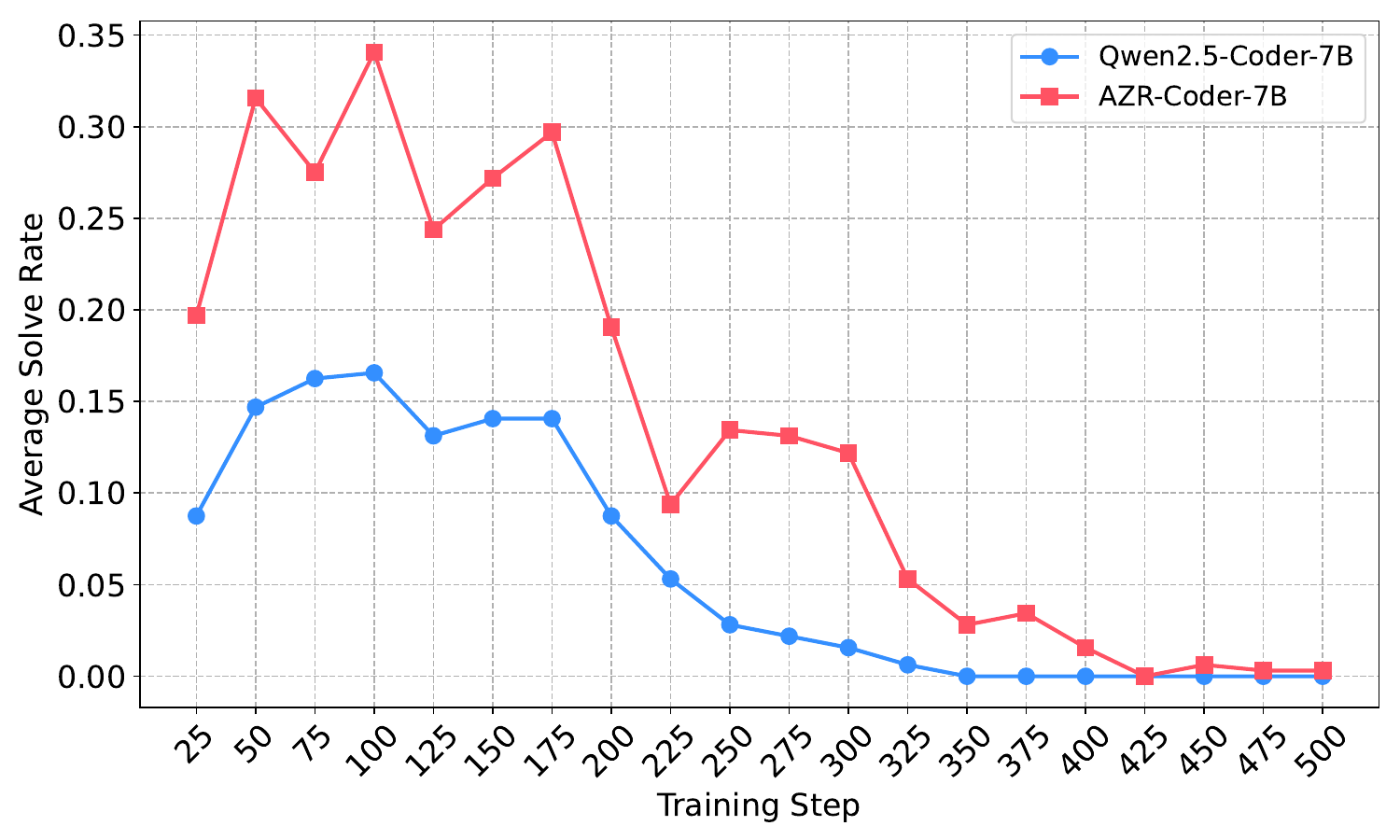}
    \end{subfigure}
    \caption{\textbf{\textsc{AZR-Coder-7B} adapts response length to question difficulty while \textsc{Qwen2.5-Coder-7B} does so at a lesser scale.} (Left) Average response length at every 25th iteration. (Right) Average solve rate at every 25th iteration.}
    \label{fig:question_difficulty}
\end{wrapfigure}

\textbf{Setup.} To study how question difficulty evolves, we created a balanced dataset of 800 deductive questions sampled from different training iterations. For each question, we generated 8 responses using \textsc{AZR-Coder-7B} and \textsc{Qwen2.5-Coder-7B} and measured the average response length and solve rate (denoted $\bar{r}_{\text{solve}}$ in Equation~\ref{eq:proposerReward}).

\textbf{Results.} As shown in Figure~\ref{fig:question_difficulty}, the proposer generates more difficult questions over time, which in turn elicit longer responses, a trait associated with improved reasoning~\citep{deepseekai2025deepseekr1incentivizingreasoningcapability, chen2025towards}. AZR produces shorter responses for easier questions, consistent with the idea of an optimal response length that avoids under or overthinking~\citep{wu2025more, su2025between}. This suggests self-play may teach the model to adjust its response length based on the perceived difficulty of a question.

\begin{tcolorbox}[colback=gray!10, colframe=black, title=Takeaway 2]
The proposer in AZR self-play generates increasingly difficult questions, and AZR appears to adapt response lengths accordingly. This implicit sensitivity to difficulty may help the model avoid both underthinking and overthinking.
\end{tcolorbox}

%% file: sections/entropy.tex
\subsection{RQ3: Entropy Collapse}

\begin{figure}[b]
    \centering
    \includegraphics[width=0.6\textwidth]{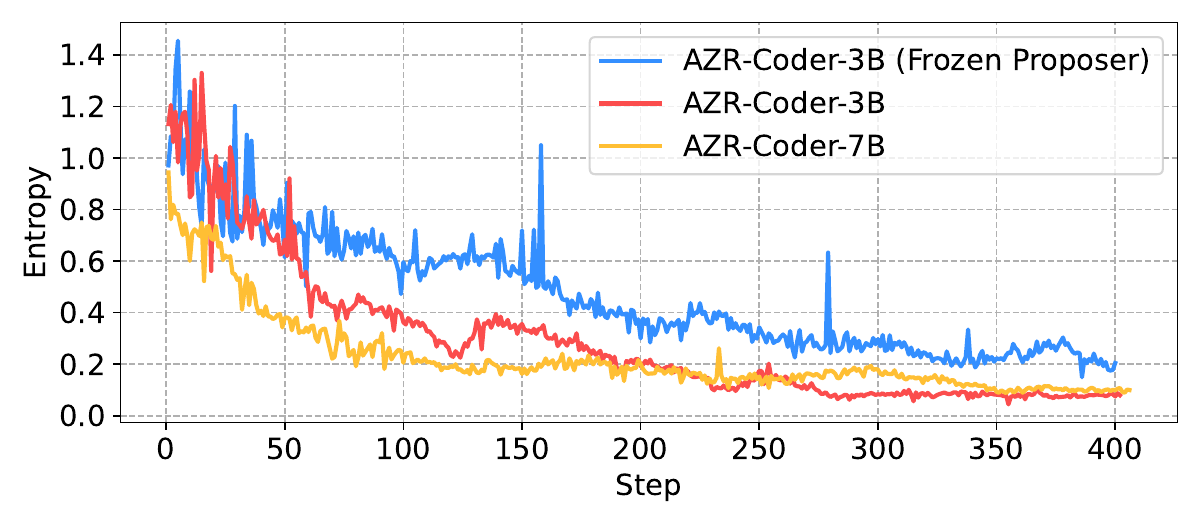}
    \caption{\textbf{Policy entropy decays at different rates based on model size and setup.} Policy entropy curves for \textsc{AZR-Coder-3B}, \textsc{AZR-Coder-3B} with a frozen proposer, and \textsc{AZR-Coder-7B}.}
    \label{fig:entropy}
\end{figure}

\textbf{Setup.}
Prior work has shown that RLVR post-training causes an \textit{entropy collapse}, where a model effectively trades exploration for validation accuracy~\citep{cui2025entropymechanismreinforcementlearning}. We examine whether self-play exhibits a similar dynamic by tracking policy entropy, defined as:
\begin{equation}\label{eq:policyEntropy}
    H(\pi_{\theta}, D) = -\mathbb{E}_{x \sim D, y \sim \pi_{\theta}}[\log \pi_{\theta}(y_t | y_{<t}, x)]
    = -\frac{1}{|D|}\sum_{x \in D}\frac{1}{|y_x|}\sum_{t=1}^{|y_x|} \log \pi_{\theta}(y_t | y_{<t}, x)
\end{equation}

\begin{wrapfigure}{r}{0.5\textwidth}
    \centering
    \vspace{-10pt} % Optional: Pulls the figure up a bit if there's too much space
    \begin{subfigure}[b]{0.24\textwidth}
        \includegraphics[width=\textwidth]{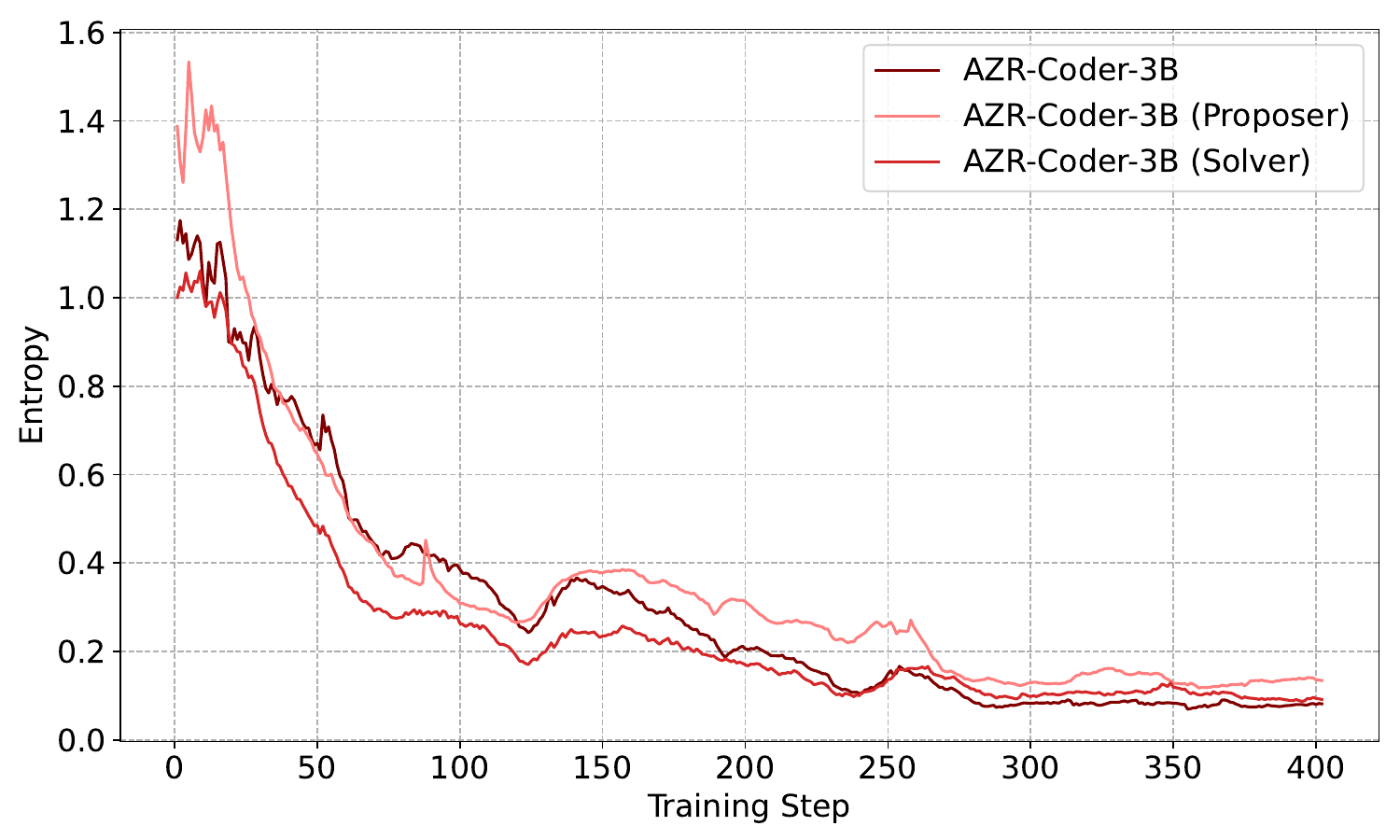}
    \end{subfigure}
    \hfill % Adds flexible space between the two figures
    \begin{subfigure}[b]{0.24\textwidth}
        \includegraphics[width=\textwidth]{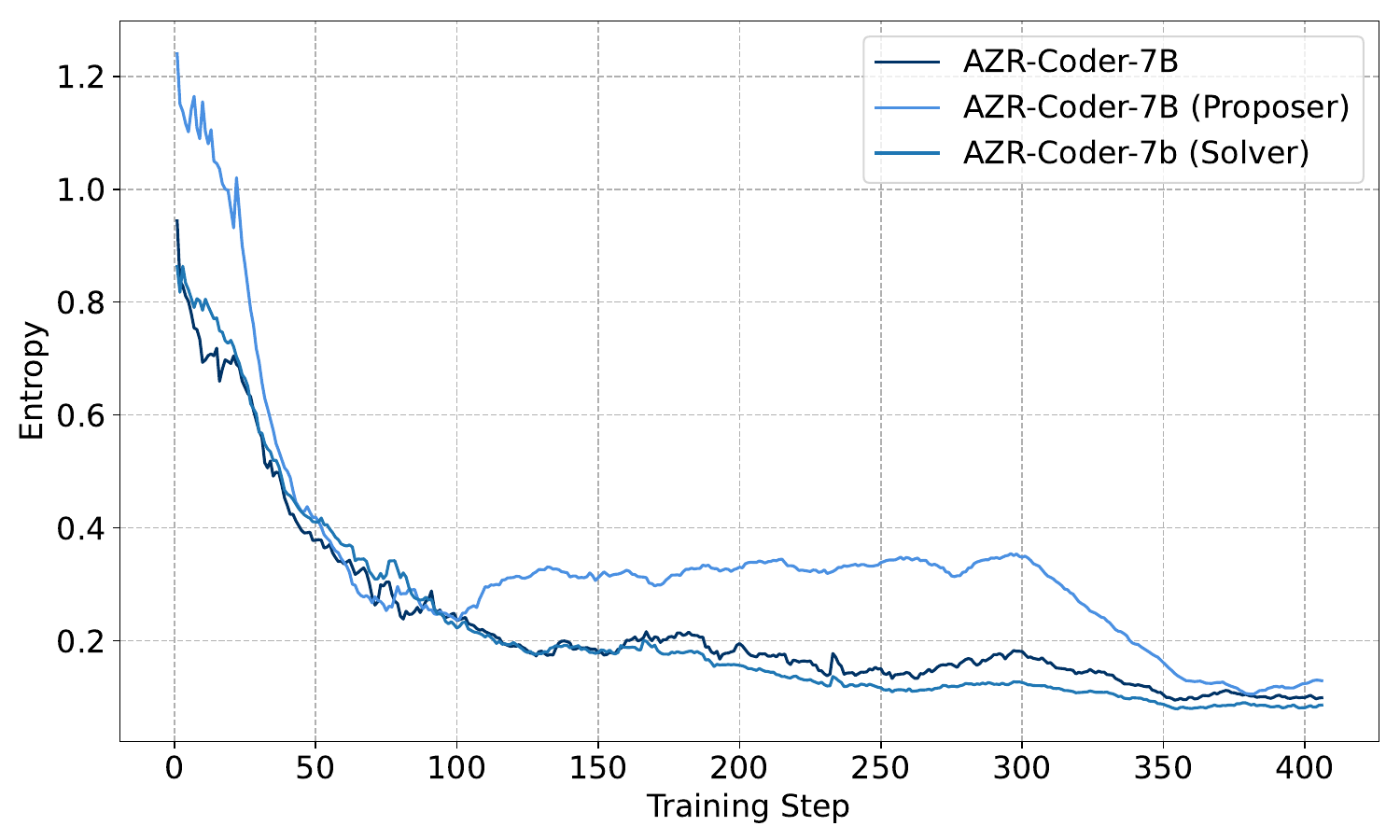}
    \end{subfigure}
    \caption{\textbf{Proposer entropy stays higher than solver entropy.} Proposer, solver, and policy entropy curves for \textsc{AZR-Coder-3B} (left) and \textsc{AZR-Coder-7B} (right)}
    \label{fig:proposer_solver_entropy}
\end{wrapfigure}

\textbf{Results.}
As shown in Figure~\ref{fig:entropy}, self-play also leads to entropy collapse, with the decay rate depending on model size and setup. The frozen-proposer variant (no PPO updates in the proposer role) maintains higher entropy than the standard 3B model, likely because in standard self-play, both proposer and solver receive gradient updates, doubling optimization pressure and accelerating collapse. When decomposed by role (Figure~\ref{fig:proposer_solver_entropy}), proposer entropy consistently remains higher than solver entropy. Although overall entropy decays, this suggests that encouraging proposer diversity could increase entropy and improve performance.

\begin{tcolorbox}[colback=gray!10, colframe=black, title=Takeaway 3]
Self-play with AZR exhibits entropy collapse, with decay rates varying by model size and proposer setup. Sustaining exploration may require explicit entropy regularization or mechanisms for promoting diverse proposer outputs.
\end{tcolorbox}

%% file: sections/sparsity.tex
\subsection{RQ4: Parameter Update Sparsity} 
\label{subsec:sparsity}

\begin{wrapfigure}{r}{0.5\textwidth}
    \vspace{-15pt}
    \centering
    \begin{subfigure}[b]{0.24\textwidth}
        \includegraphics[width=\textwidth]{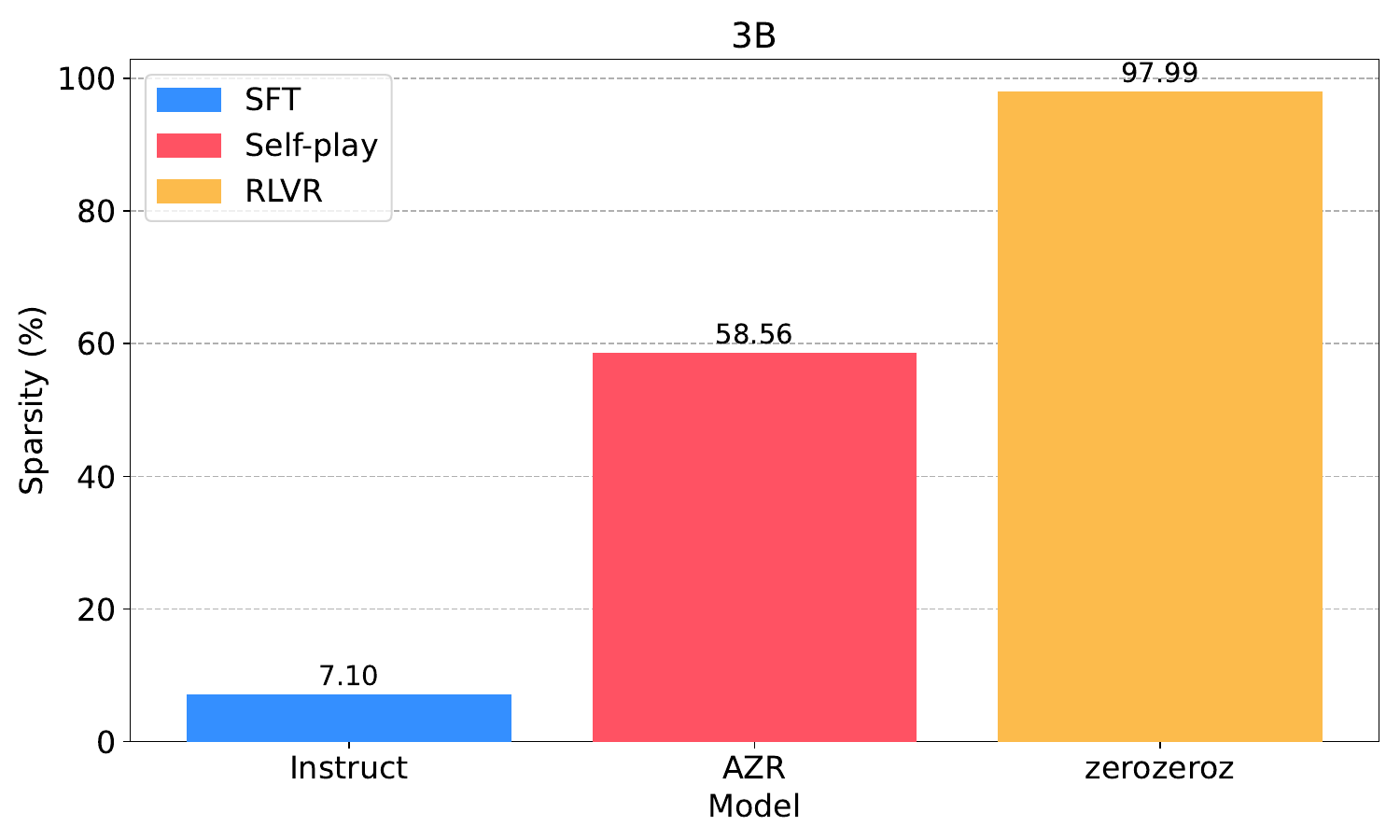}
    \end{subfigure}
    \begin{subfigure}[b]{0.24\textwidth}
        \includegraphics[width=\textwidth]{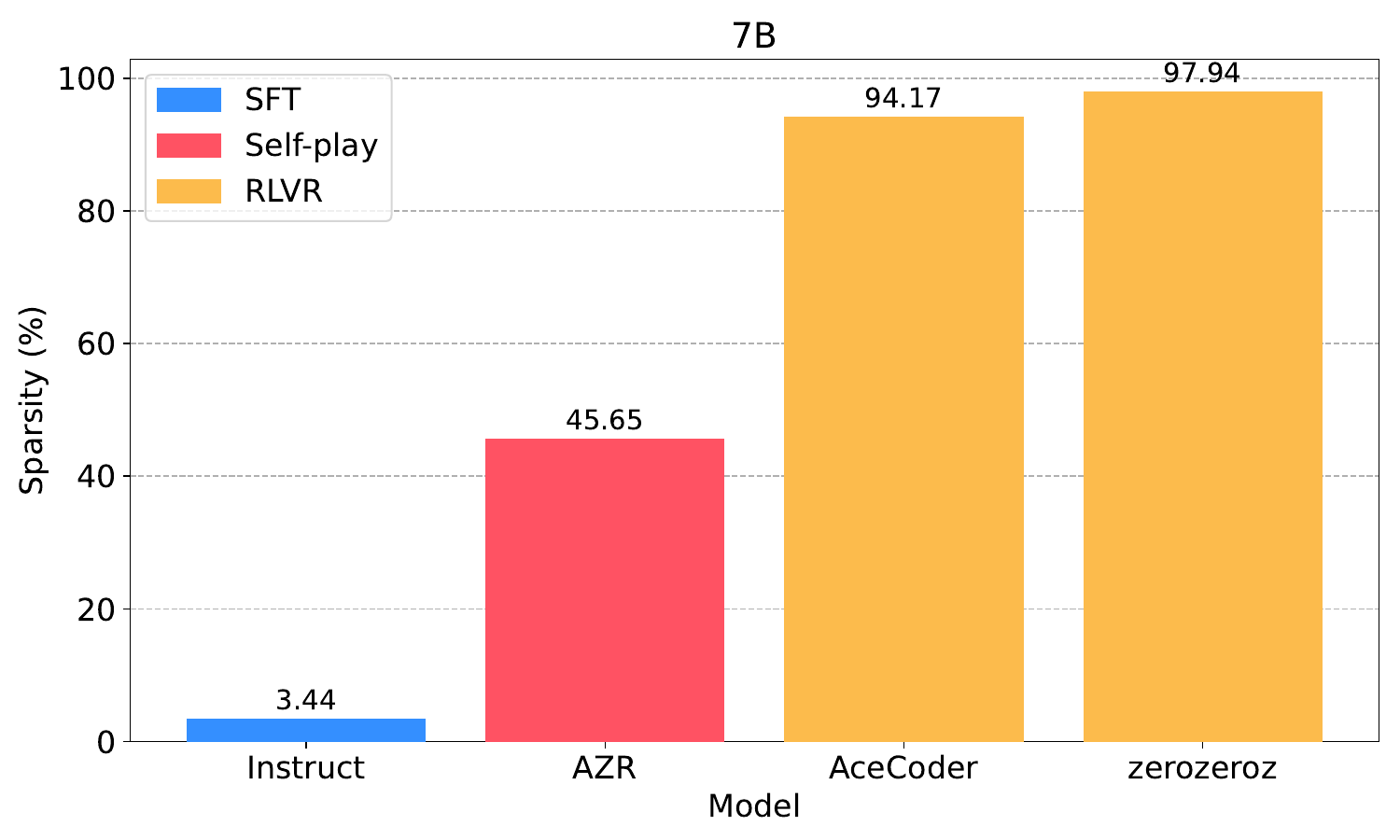}
    \end{subfigure}
    \caption{\textbf{Self-play has distinct update sparsity compared to RL-tuned and SFT models.} Update sparsity comparison between public checkpoints of fine-tuned models and their corresponding base models: (left) \textsc{Qwen2.5-Coder-3B} and (right) \textsc{Qwen2.5-Coder-7B}.}
    \label{fig:sparsity} 
\end{wrapfigure}

\textbf{Setup.}
Following \citet{mukherjee2025reinforcement}, who showed that supervised fine-tuning (SFT) produces dense parameter updates while reinforcement learning (RL) is sparse, we analyze the \textbf{update sparsity} of self-play. This metric measures the proportion of parameters unchanged between two checkpoints:
\begin{equation}\label{eq:updateSparsity}
    S(\theta^0, \theta^1) \coloneq 1 - \frac{\lVert\theta^1 - \theta^0\rVert_0}{n}, 
\end{equation}
where $\lVert\cdot\rVert_0$ counts non-zero elements. We compare our self-play models (\textsc{AZR-Coder}) against public SFT and RLVR checkpoints that use the same base model.

% \textbf{Results.}
% Figure~\ref{fig:sparsity} shows that self-play training results in an intermediate level of parameter update sparsity, positioned between the dense updates from SFT and the extremely sparse updates from RLVR. For instance, \textsc{AZR-Coder-7B} reaches about $45.7\%$ sparsity. \citet{mukherjee2025reinforcement} suggested that sparsity reflects training on in-distribution data; thus, the intermediate sparsity of our model may arise from its dual process of both generating novel data and subsequently solving it.

\textbf{Results.}
Figure~\ref{fig:sparsity} shows that self-play produces an intermediate level of parameter update sparsity, denser than RLVR but sparser than SFT. For example, \textsc{AZR-Coder-7B} reaches about $45.7\%$ sparsity. This pattern likely reflects the dual nature of self-play, where the model both generates new data and learns from solving it, leading to updates that are partly in-distribution yet still exploratory.

\begin{tcolorbox}[colback=gray!10, colframe=black, title=Takeaway 4]
AZR self-play leads to intermediate update sparsity: denser than RLVR but sparser than SFT, reflecting its dual role of data generation and solution.
\end{tcolorbox}

%% file: sections/reward.tex
\subsection{RQ5: Proposer Reward Function}\label{sec:reward}

\textbf{Setup.}
Prior work suggests that a solve rate near $0.5$ yields strong gradient signals in RLVR with GRPO-style advantage calculations~\citep{shi_efficient_2025, yu2025dapo}. Building on this, we test a modified reward to explicitly encourage the proposer to generate questions of this difficulty:

\begin{equation} \label{eq:0.5_peak}
    r_{\text{propose}} =
    \begin{cases}
        0 & \text{if } \bar{r}_{\text{solve}} \in \{0, 1\}, \\
        1 - 2 \, \big|\bar{r}_{\text{solve}} - 0.5\big| & \text{otherwise}.
    \end{cases}
\end{equation}
This function gives the maximum reward when the average solve rate, $\bar{r}_{\text{solve}}$, is $0.5$, incentivizing the proposer to generate questions that are neither too easy nor too hard.

\begin{figure}[t]
    \centering
    \begin{subfigure}[b]{0.4\textwidth}
        \includegraphics[width=\textwidth]{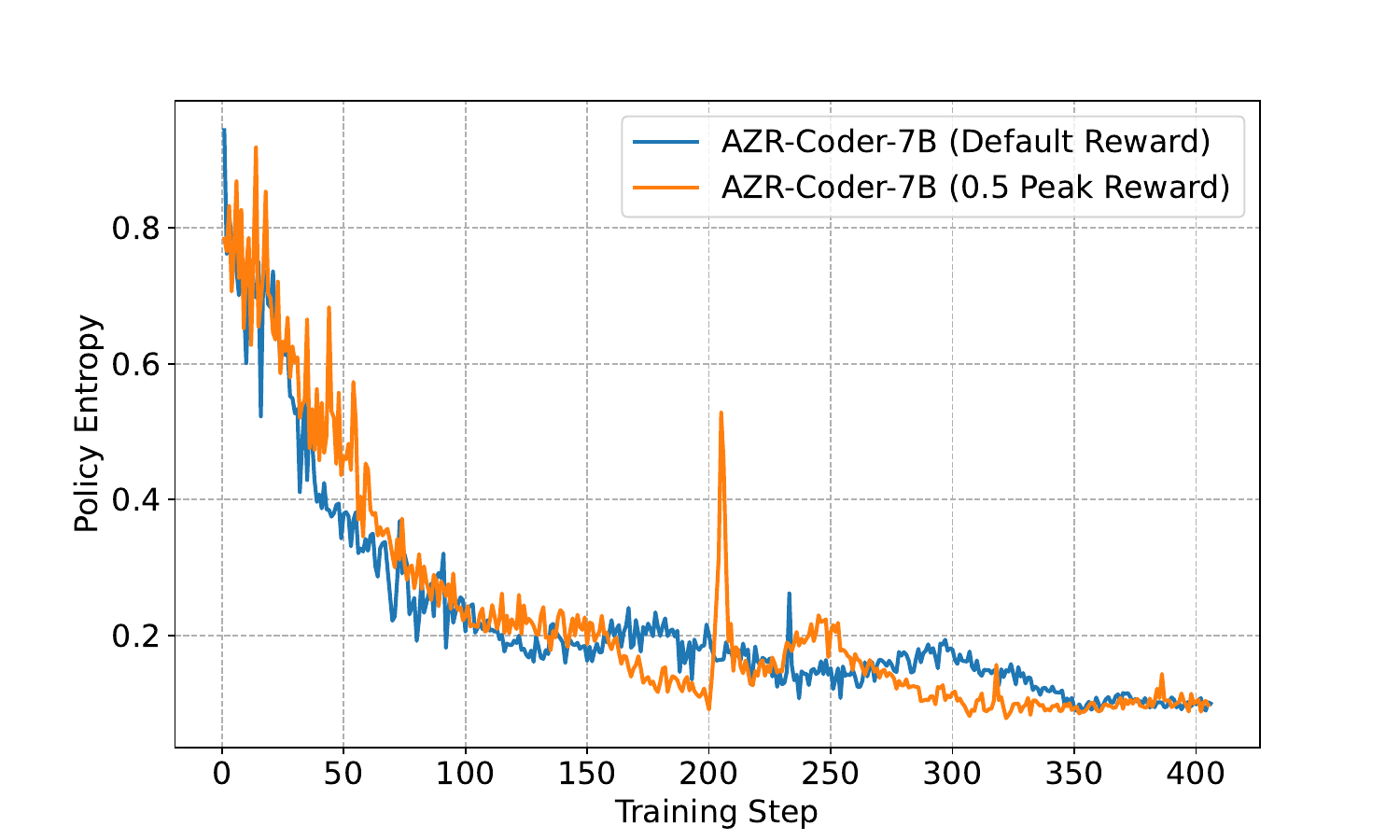}
    \end{subfigure}
    \begin{subfigure}[b]{0.36\textwidth}
        \includegraphics[width=\textwidth]{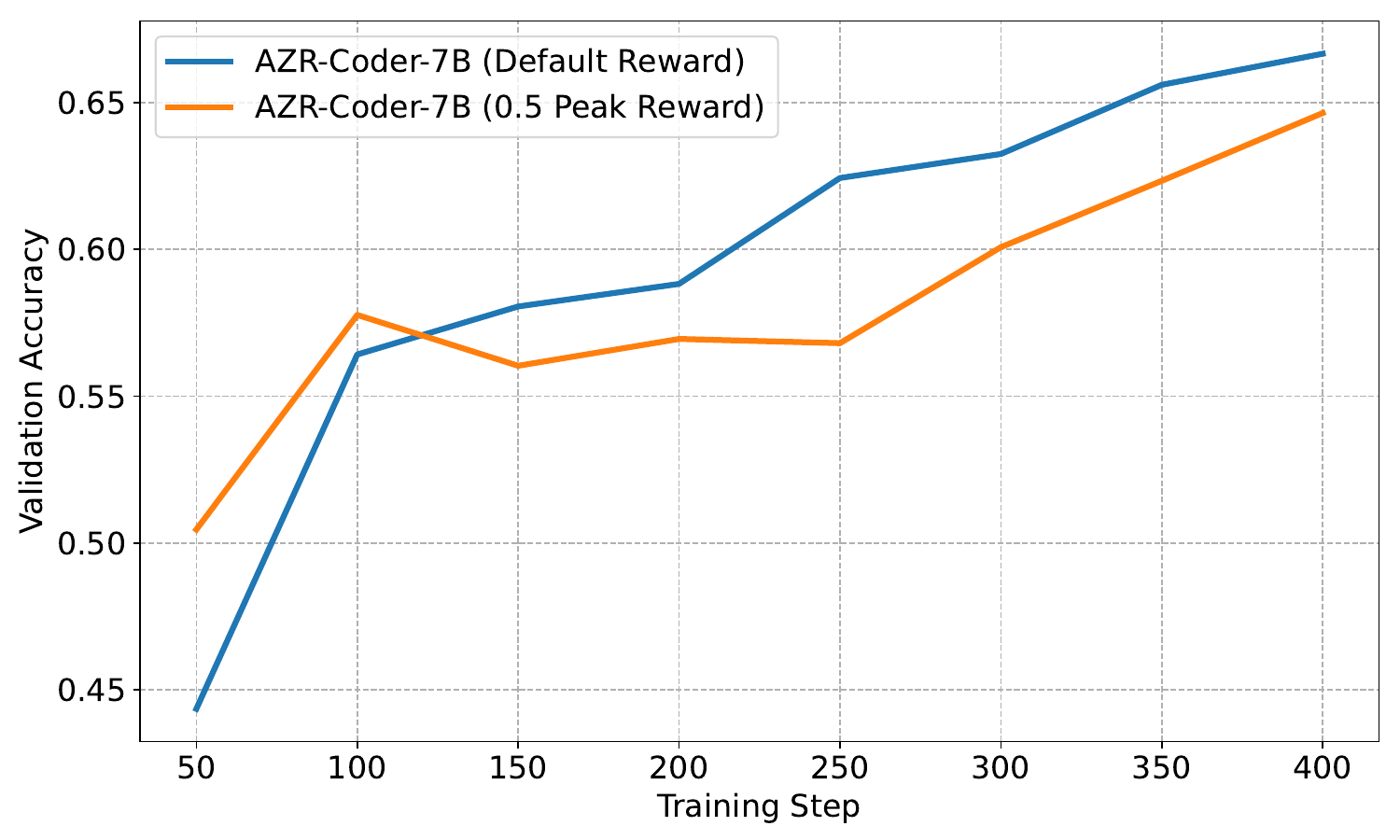}
    \end{subfigure}
    \caption{\textbf{Training dynamics of different proposer rewards.} (Left) Policy entropy training dynamics. (Right) Validation accuracy training dynamics}
    \label{fig:ent_acc_reward} 
\end{figure}

\textbf{Results.}
As shown in Figure~\ref{fig:ent_acc_reward}, this reward function had little effect on entropy dynamics and reduced final validation accuracy by 2\%. While \citet{huang_r-zero_2025} used it successfully in \emph{R-Zero}, our results suggest their gains stem from other system components.

\begin{tcolorbox}[colback=gray!10, colframe=black, title=Takeaway 5]
Altering the proposer’s reward to target a 50\% solve rate does not improve AZR performance and slightly reduces validation accuracy. This suggests that reward shaping alone is insufficient, with other components playing a larger role in AZR’s self-play gains.
\end{tcolorbox}

\section{Limitations \& Future Work}

Our study is limited in scope to a single self-play framework (AZR) and two model sizes (3B and 7B). The patterns we observe with AZR may not hold across other self-play frameworks or larger models~\cite{huang_r-zero_2025,liu2025spiral}. Nevertheless, given the highly similar proposer–solver structures that most recent self-play methods adopt, we expect that our findings are broadly informative.  

% There are several promising directions for future work. First, self-play remains bounded by the reasoning capacity of the base model, so approaches that can inject probability mass into novel reasoning trajectories or otherwise expand reasoning capacity in RL-tuned models are of particular interest. Second, automatic curriculum learning through self-play may elicit new capabilities if designed effectively. Third, given the unique parameter update sparsity observed in self-play, it is important to investigate whether self-play models suffer from catastrophic forgetting. Prior work has shown that RL post-training tends to reduce forgetting relative to SFT \citep{shenfeld2025rl}, and \citet{mukherjee2025reinforcement} suggest that this effect may be linked to update sparsity. Assessing whether self-play shares this property is therefore an important open question. Finally, preventing entropy collapse remains a critical challenge. Future work could explore proposer modifications or explicit entropy regularization to sustain exploration during training. 

There remain several promising directions for future research. First, self-play is still bounded by the base model's reasoning capacity, motivating approaches that can expand reasoning support or inject probability mass into novel trajectories. Second, automatic curriculum learning within self-play could elicit new capabilities if designed effectively. Third, given the distinctive parameter update sparsity observed, it is worth investigating whether self-play mitigates or exacerbates catastrophic forgetting, an effect linked to sparsity in prior RL studies~\citep{shenfeld2025rl,mukherjee2025reinforcement}. Finally, preventing entropy collapse remains a key challenge; future work could explore proposer modifications or explicit entropy regularization to sustain exploration during training.

%% file: sections/conclusion.tex
% \section{Conclusion}
% We analyzed self-play for LLM reasoning as implemented by AZR and characterized it through its training dynamics and differences with the base model. We showed that AZR's reasoning capacity is still bounded by the base model. However, the proposer component may drive improved pass$@k$ performance at large $k$ through question diversity. We further highlight the importance of the proposer through our other experiments. The proposer is able to create a curriculum increasing in difficulty throughout training. Using the questions created by the proposer, we evaluate AZR and its base model and find that AZR is able to associate response length with question difficulty, potentially preventing over and underthinking through co-evolution with the solver. We also show that improvements to the proposer component of self-play is likely the best option in preventing entropy collapse in our entropy experiments. The most distinctive feature of self-play was its parameter update sparsity, leading to further questions as to what drives this difference with RL-tuned and SFT models and whether self-play experiences catastrophic forgetting. Finally, we attempt to improve the proposer by altering the reward function to encourage questions that lead to strong gradient signals by targeting 50\% solve rate. However, we find that improving the proposer will take something much more explicit. 

\section{Conclusion}
% We analyzed self-play for LLM reasoning using AZR, focusing on its training dynamics and limits relative to the base model. AZR’s reasoning capacity remains bounded by the base, but the proposer plays a key role. It improves pass$@k$ at larger $k$ through question diversity, generates a curriculum of increasing difficulty, and adapts response length to question difficulty. We also find that entropy collapse persists, pointing to the proposer as the most promising target for improvement. Finally, self-play shows distinctive parameter update sparsity, raising questions about its relation to catastrophic forgetting and how proposer rewards can be refined to provide stronger learning signals.
Our analysis of the self-play framework AZR reveals that while its reasoning capacity is bounded by the base model, the proposer is the critical component for improvement. The proposer drives performance by generating a diverse and progressively difficult curriculum of questions. We find that self-play exhibits a unique parameter update sparsity and still undergoes entropy collapse, pointing to the proposer as the most promising target for future work. Finally, self-play shows distinctive parameter update sparsity, raising questions about its relation to catastrophic forgetting and how proposer rewards can be refined to provide stronger learning signals.

\vspace{-0.5em}

\section*{Acknowledgment}

This material is based upon work supported by the National
Science Foundation Award \#2447631. 

%% file: appendix.tex
\appendix
\input{sections/AZR}
\input{sections/related}

\input{sections/limitations}
\input{sections/invisible_proof}
\section{Response Length} 
The following is the average response length (number of tokens in a response) for each iteration across the three tasks and two roles. 
\begin{figure}[th]
    \centering
    \begin{subfigure}[b]{0.4\textwidth}
        \includegraphics[width=\textwidth]{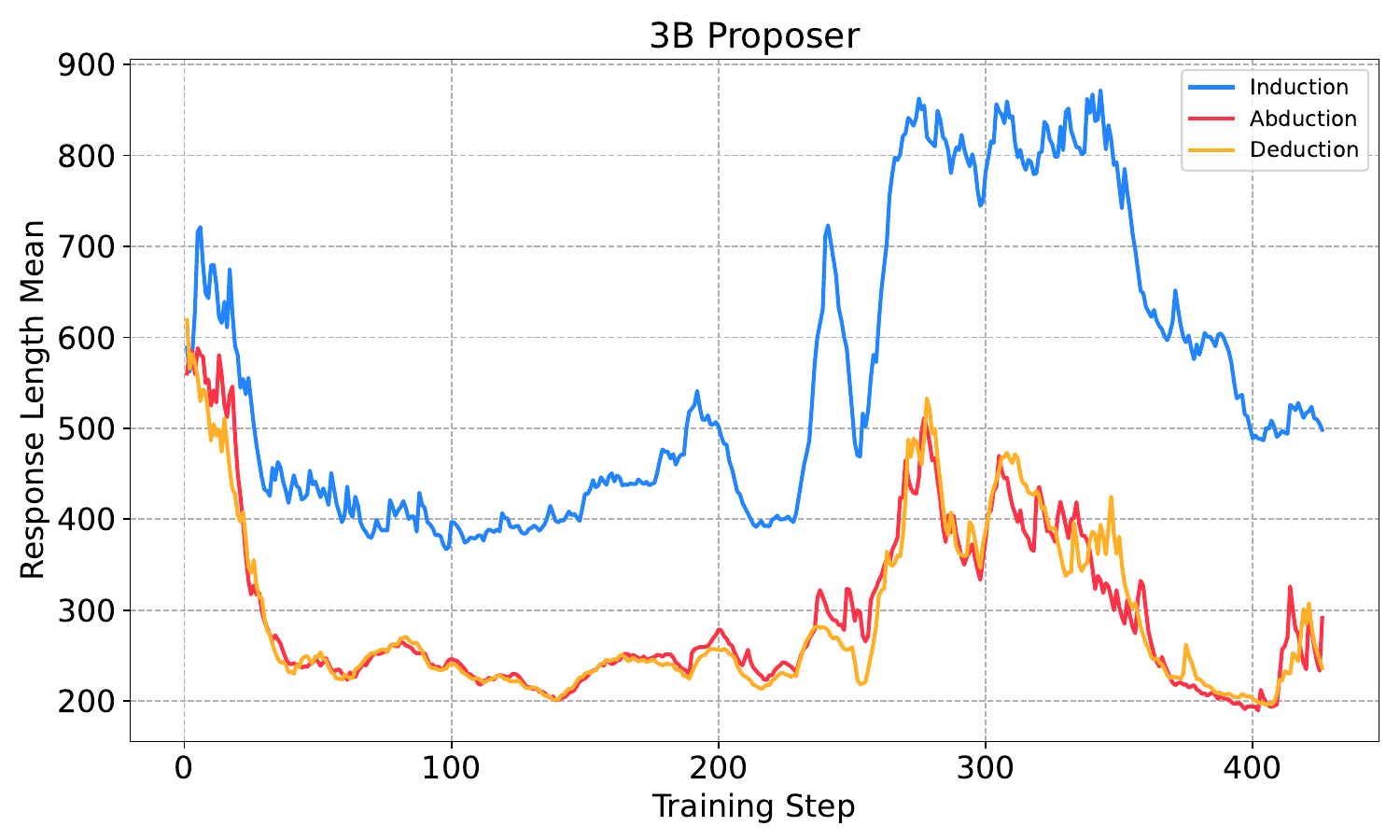}
    \end{subfigure}
    \begin{subfigure}[b]{0.4\textwidth}
        \includegraphics[width=\textwidth]{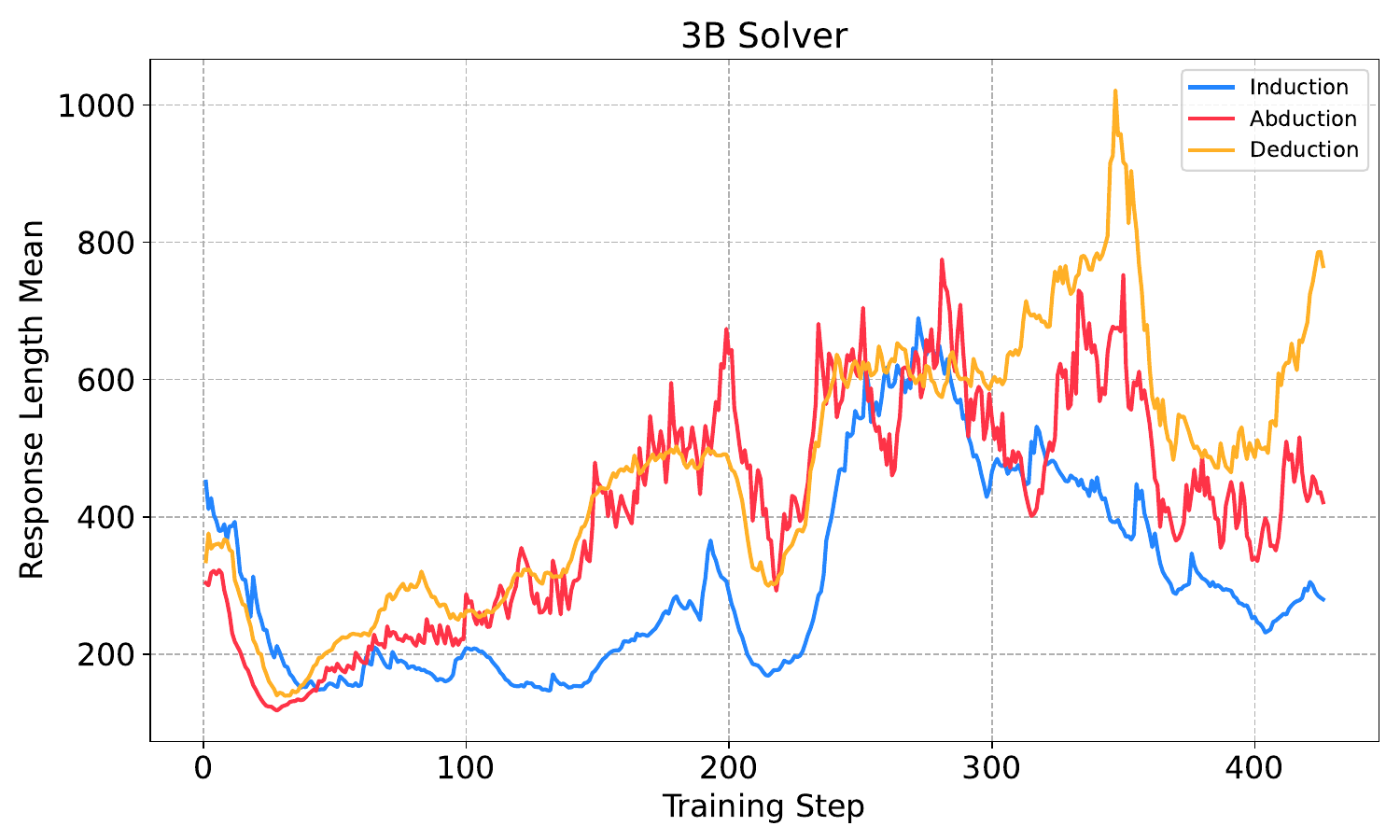}
    \end{subfigure} \\
    \begin{subfigure}[b]{0.4\textwidth}
        \includegraphics[width=\textwidth]{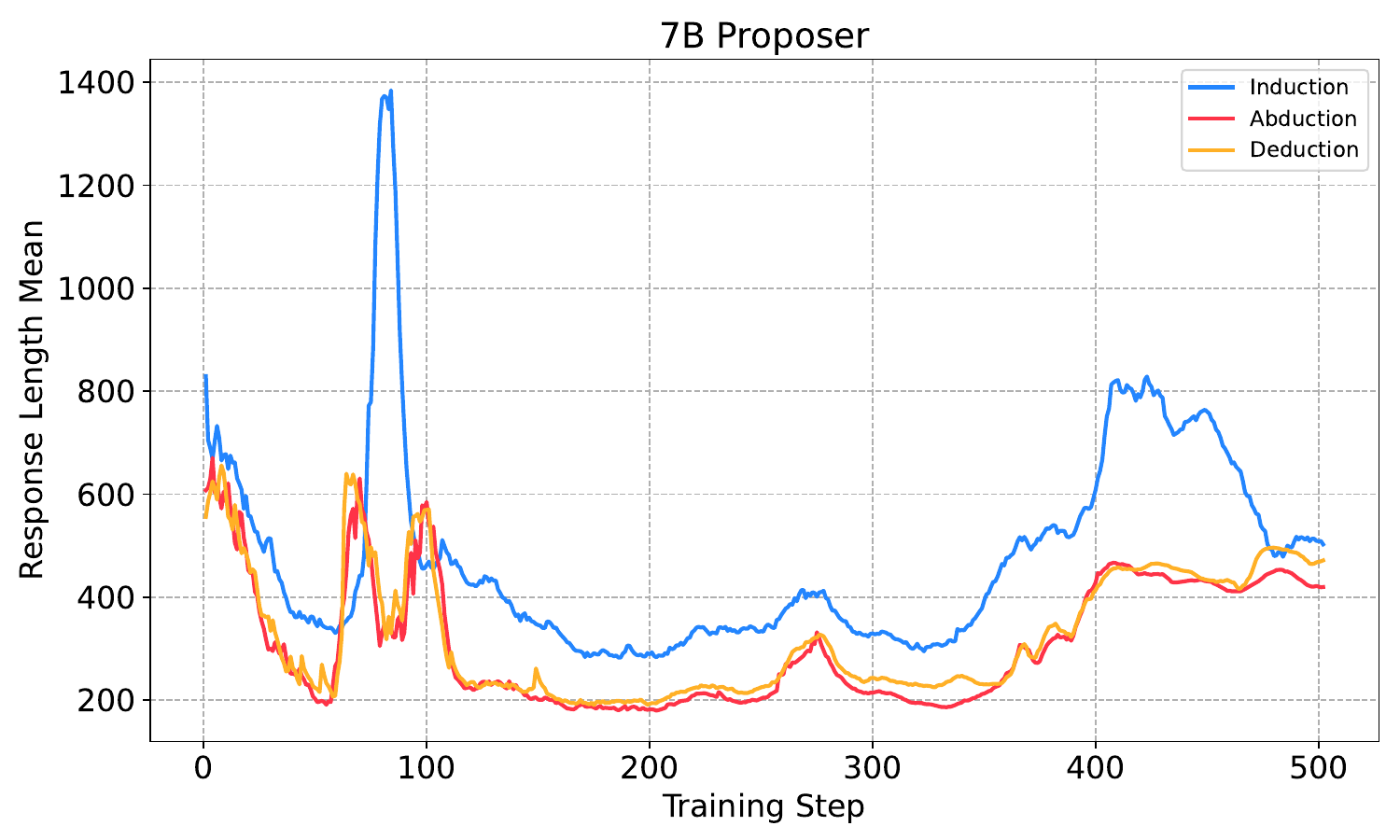}
    \end{subfigure}
    \begin{subfigure}[b]{0.4\textwidth}
        \includegraphics[width=\textwidth]{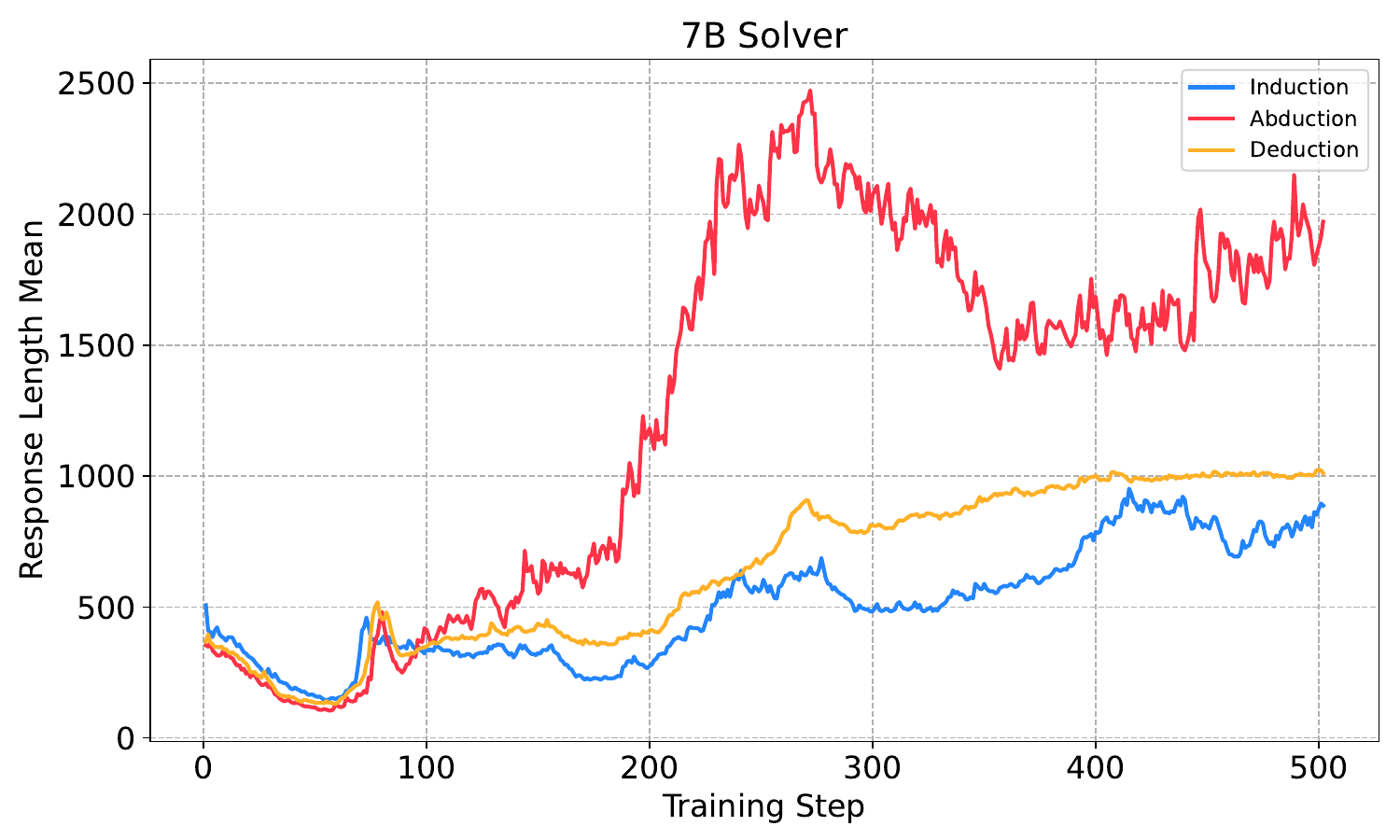}
    \end{subfigure}
    \caption{Response length mean per training step exponentially smoothed across model sizes and roles}
    \label{fig:response_lengths} 
\end{figure}

%% file: sections/AZR.tex
\section{Absolute Zero Reasoner}
\label{sec:AZR}

\paragraph{Rewards.}
AZR assigns complementary rewards for the proposer and solver~\cite{zhao2025absolute}. The proposer is incentivized to create tasks of moderate difficulty using a learnability reward based on the solver’s average success rate. If a task is always solved or never solved, the proposer receives no reward, while partially solvable tasks yield a higher reward:  

\begin{equation}\label{eq:proposerReward}
    r_{\text{propose}} = 
    \begin{cases}
        0 & \text{if } \bar{r}_{\text{solve}} = 0 \text{ or } \bar{r}_{\text{solve}} = 1 \\
        1 - \bar{r}_{\text{solve}} & \text{otherwise}
    \end{cases}
\end{equation}

The solver receives a binary correctness reward for producing the ground-truth answer, where equality is evaluated in Python:  

\begin{equation}\label{eq:solverReward}
    r_{\text{solve}} = \mathbb{I}_{(y = y^\star)}
\end{equation}

Both roles are further regularized with a format-aware penalty, ensuring adherence to the required \texttt{<think>} and \texttt{<answer>} structure~\cite{deepseekai2025deepseekr1incentivizingreasoningcapability}:  

\begin{equation}\label{overallReward}
    R = 
    \begin{cases}
        r_{\text{role}} & \text{if the completion is passable} \\
        -0.5 & \text{if the completion is wrong but well formatted} \\
        -1 & \text{if the completion has formatting errors}
    \end{cases}
\end{equation}

\paragraph{Learning Different Modes of Reasoning.}
AZR implements three reasoning modes over a coding triplet $(p,i,o)$ consisting of a program, input, and output~\cite{zhao2025absolute}. In deduction, the model predicts $o$ from $(p,i)$, with outputs verified by type-aware equality. In abduction, it infers a plausible input $i$ from $(p,o)$, rewarding solutions that reproduce the correct output even when programs are non-bijective. In induction, it synthesizes a program $p$ from partial input–output pairs, with held-out examples used to ensure generalization beyond memorization. Together, these modes enable reasoning across different components of the triplet while using code execution as both an expressive interface and a verifiable environment.

% \subsection{Algorithm}
% Each training iteration, the proposer creates a batch for each reasoning type. Then, the solver answers each prompt in the batch. Finally, the rewards are generated and used to update the proposer and solver using \emph{Task-Relative Reinforce++}, which is simply the Reinforce++ algorithm with batch level statistics for each reasoning type and role leading to 6 different baselines (3 reasoning types times 2 roles).

\paragraph{Learning Algorithm.}
Training in AZR begins by seeding task buffers with valid program–input–output triplets generated from the base model, optionally starting from a trivial “zero” triplet. Separate buffers are maintained for deduction, abduction, and induction, and each proposer samples past triplets as in-context references to generate new tasks. Validity is enforced through a lightweight pipeline that executes candidate programs, checks syntax and safety, and restricts to deterministic outputs. To ensure stable training, if insufficient valid tasks are generated in a batch, examples are backfilled from the existing buffers.  

Once validated, tasks are presented to the solver in role-specific forms, and solutions are verified by equality checks appropriate to deduction, abduction, or induction. Rewards are then assigned to both proposer and solver, and model parameters are updated with Task-Relative REINFORCE++ (TRR++), which maintains separate baselines for each task–role combination. This structure reduces variance across six training configurations and enables AZR to expand its curriculum and improve through self-play.

%% file: sections/related.tex
\section{Related Works} %move to appendix?

\subsection{Reinforcement Learning with Verifiable Rewards}

A central advance in RLVR has been the development of critic-free algorithms such as Group Relative Policy Optimization (GRPO), which reduced the computational cost of reinforcement learning for LLM post-training~\citep {deepseekai2025deepseekr1incentivizingreasoningcapability}. Building on this foundation, many works have refined GRPO and studied its limitations. 

For example, \citet{liu_understanding_2025} identified bias in the advantage calculation of GRPO and proposed removing the group standard deviation term. \citet{yu2025dapo} introduced DAPO, which incorporates techniques such as dynamic sampling, filtering out overly easy or difficult samples that provide little gradient signal, and a clip higher method to encourage higher entropy. Acting as a broad empirical study, \citet{liu_part_2025} demonstrated that simple design choices such as combining group-level mean with batch-level variance for advantage estimation, together with token-level loss aggregation, are sufficient to outperform both GRPO and more complex variants like DAPO.

Many works have also investigated RLVR from an entropy perspective as well. \citet{cui2025entropymechanismreinforcementlearning} observed that tokens with high covariance (high probability and advantage) drive entropy collapse. To address high covariance tokens, they either clipped those tokens from the gradient calculation or applied a KL penalty, preventing entropy collapse \cite{cui2025entropymechanismreinforcementlearning}. \citet{wang2025beyond} also clips tokens from gradient calculation using token-level entropy as the determiner and only training on the 20\% highest entropy tokens, observing that low entropy tokens contribute to entropy collapse. \citet{cheng2025reasoning} also observes that high entropy tokens are important for performance gains and exploration, and they propose an entropy aware advantage term, leading to pass$@k$ gains on benchmarks.  

Curriculum-based approaches have also shown promise, where the model is presented with problems matched to its current ability, for instance, those with a solve rate of about fifty percent across rollouts~\citep{shi_efficient_2025, sun_improving_2025}. 

Alongside these improvements, other studies have examined whether RLVR actually increases the reasoning ability of models. \citet{yue2025does} showed that RLVR substantially improves pass$@k$ when $k$ is small, but that for large $k$ the base model consistently surpasses its RLVR trained counterpart. Extending this finding, \citet{wu_invisible_2025} provided both theoretical and empirical evidence that standard RLVR cannot escape the reasoning capacity of the base model.

\subsection{LLMs for Mathematics}
Given its inherent verifiability, mathematics has emerged as a compelling domain for the post-training of LLMs. This has spurred a significant body of work aimed at augmenting the mathematical capabilities of these models through RLVR~\cite{shao2024deepseekmath, wang2025reinforcement, liu2025acereasonnemotron11advancingmath}. In particular, Google DeepMind's AlphaProof achieved silver-medal standing at the International Mathematical Olympiad (IMO) by leveraging self-play with a pretrained LLM~\cite{DeepMindAlphaProof2024}. This approach utilized the formal language Lean, a strategy comparable to the use of Python in the AZR framework~\cite{10.1007/978-3-030-79876-5_37}. Such findings reinforce the idea that an LLM's coding aptitude is foundational to its downstream mathematical performance after post-training~\cite{shao2024deepseekmath, zhao2025absolute}.

% \subsection{Self-play}
% Several works utilizing self-play for LLM post-training have shown its efficacy in improving LLM reasoning capabilities. Specifically, many works show that allowing the LLM to propose its own dataset while also acting as the solver can lead to surprising gains \cite{zhao2025absolute, chen_self-questioning_2025, huang_r-zero_2025}. \cite{liang_beyond_2025} utilizes self-play to evolve questions from existing datasets and shows that this leads to not only increased pass$@1$ scores but also increased pass$@k$ scores when $k$ is large. Others have also shown gains in instruction following through self-play \cite{cheng_spar_2025, dong_self-play_2024}. 

\subsection{Self-play}

Recent work has demonstrated that self-play can be an effective strategy for LLM post-training, with consistent gains in reasoning ability. A common approach is to allow the model to generate its own training data by proposing problems and then solving them, which has been shown to yield strong improvements~\citep{zhao2025absolute, chen_self-questioning_2025, huang_r-zero_2025}. \citet{liang_beyond_2025} extend this idea by evolving questions from existing datasets and report gains not only in pass$@1$ performance but also in pass$@k$ when $k$ is large. Self-play has also proven useful beyond reasoning tasks, with several studies showing improvements in instruction following and robustness~\citep{cheng_spar_2025, dong_self-play_2024}.

%% file: sections/invisible_proof.tex
\section{AZR Inherits the Invisible-Leash Support Bound} %move to appendix?
\label{sec:azr-il}

\vspace{-0.5ex}
\paragraph{Setting.}
Let $\mathcal X$ denote prompts or tasks and $\mathcal Y$ denote token sequences (solutions).
A base model $q_0(y\mid x)$ initializes both AZR roles: the proposer $\pi^{\mathrm P}_t$ and the solver $\pi^{\mathrm S}_t$. 
Each role is trained on verifiable rewards using on-policy policy-gradient updates (REINFORCE or PPO style), as in Absolute Zero (AZR) \cite{zhao2025absolute}.
The generic on-policy update for either role is
\[
\pi_{t+1}(y\mid x)
= (1-\gamma_t)\,\frac{\pi_t(y\mid x)\,w_t(y,x)}{Z_t(x)}
\;+\; \gamma_t\,\mu_t(y\mid x),
\]
where $w_t(\cdot,\cdot)\!\ge\!0$ is the clipped or exponentiated advantage weight and $Z_t$ is the normalization factor.
AZR sets $\gamma_t = 0$ (no explicit exploration) and does not use off-policy data mixing \cite{zhao2025absolute}.

\paragraph{Invisible-Leash principle (reused).}
For any on-policy reweighting of the form above with $\gamma_t=0$, \emph{support is preserved}:
if $\pi_t(y\mid x)=0$ then $\pi_{t+1}(y\mid x)=0$, since $w_t$ is only evaluated on $y\!\sim\!\pi_t(\cdot\mid x)$ and no mass is injected off-support.
By induction from $\pi_0=q_0$, $\mathrm{supp}(\pi_t(\cdot\mid x))\subseteq \mathrm{supp}(q_0(\cdot\mid x))$ for all $t$ \cite{wu_invisible_2025}.

\begin{theorem}[Role-wise support preservation for AZR]
\label{thm:azr-support}
Initialize with $\pi^{\mathrm P}_0=\pi^{\mathrm S}_0=q_0$.
Under AZR's on-policy updates with $\gamma_t\equiv 0$ and no off-policy data,
\[
\mathrm{supp}\!\left(\pi^{\mathrm P}_t(\cdot\mid z)\right)\subseteq \mathrm{supp}\!\left(q_0(\cdot\mid z)\right),
\qquad
\mathrm{supp}\!\left(\pi^{\mathrm S}_t(\cdot\mid x)\right)\subseteq \mathrm{supp}\!\left(q_0(\cdot\mid x)\right)
\quad \forall t.
\]
\end{theorem}

\noindent\emph{Proof sketch.}
Apply the Invisible-Leash support argument \cite{wu_invisible_2025} separately to the proposer and the solver.
Both roles update via on-policy tilting (no mixing), so each preserves its support.
Endogenous task selection in AZR \cite{zhao2025absolute} does not alter the conditional support property of $\pi^{\mathrm S}_t(\cdot\mid x)$; it depends only on the solver’s update rule. \qed

\begin{corollary}[Reasoning boundary / zero-probability barrier]
\label{cor:zero-barrier}
Let $A_x := \{y\in\mathcal Y:\text{verifier accepts }(x,y)\}$.
If $q_0(A_x)=0$, then for all AZR iterates $t$ and $k\ge 1$,
\[
\Pr\!\left[\exists\,\text{success in }k\text{ i.i.d.\ samples from }\pi^{\mathrm S}_t(\cdot\mid x)\right]=0.
\]
Thus AZR cannot produce verifiably correct sequences for tasks whose correct solutions have zero probability under the base model \cite{wu_invisible_2025}.
\end{corollary}

\paragraph{Empirical-support variant (finite-precision LLMs).}
Define $\mathrm{supp}_\varepsilon(q_0):=\{y:q_0(y\mid x)>\varepsilon\}$ and $S_\varepsilon:=\mathcal Y\!\setminus\!\mathrm{supp}_\varepsilon(q_0)$.
With any trust-region step (e.g., PPO clipping or per-step KL $\le\delta$), the Invisible-Leash reweighting bound yields
\[
\pi^{\mathrm S}_{t+1}\!\big(S_\varepsilon\mid x\big)
\;\le\; C(\delta)\cdot \pi^{\mathrm S}_{t}\!\big(S_\varepsilon\mid x\big),
\]
so mass outside the base model’s $\varepsilon$-support remains negligible absent explicit exploration ($\gamma_t\!>\!0$) or off-policy data \cite{wu_invisible_2025}.
AZR uses $\gamma_t\!=\!0$ and on-policy PPO, hence inherits this empirical-support limitation \cite{zhao2025absolute}.

\paragraph{When can AZR escape the bound?}
Only by injecting off-support mass (e.g., $\gamma_t>0$ with suitably covering $\mu_t$), adding off-policy data, changing the base model (capacity/tools), or otherwise breaking on-policy tilting \cite{wu_invisible_2025}.